\documentclass{article}

\usepackage{arxiv}

\usepackage[utf8]{inputenc} % allow utf-8 input
\usepackage[T1]{fontenc}    % use 8-bit T1 fonts
\usepackage{hyperref}       % hyperlinks
\usepackage{url}            % simple URL typesetting
\usepackage{booktabs}       % professional-quality tables
\usepackage{amsfonts}       % blackboard math symbols
\usepackage{nicefrac}       % compact symbols for 1/2, etc.
\usepackage{microtype}      % microtypography
\usepackage{cleveref}       % smart cross-referencing
\usepackage{lipsum}         % Can be removed after putting your text content
\usepackage{graphicx}
\usepackage{doi}
\usepackage[authoryear]{natbib} % 设置为作者-年份引用格式

\title{Optimization of Transformer heart disease prediction model based on particle swarm optimization algorithm}

\date{}

\newif\ifuniqueAffiliation
\uniqueAffiliationtrue

\ifuniqueAffiliation % Standard variant of author block
\author{ {\hspace{1mm}Jingyuan Yi$^{1*}$} \\
	Information Networking Institute\\
	Carnegie Mellon University\\
	Pittsburgh, PA, 15213 \\
	\texttt{jingyuay@alumni.cmu.edu} \\
	\And
    {\hspace{1mm}Peiyang Yu$^{1*}$} \\
	Information Networking Institute\\
	Carnegie Mellon University\\
	Pittsburgh, PA, 15213 \\
	\texttt{peiyangy@alumni.cmu.edu} \\
	\And
	{\hspace{1mm}Tianyi Huang$^{2}$} \\
	Department of Electrical Engineering \& Computer Sciences\\
	University of California, Berkeley\\
	Berkeley, CA, 94720 \\
	\texttt{tianyihuang@berkeley.edu} \\
	\And  
	{\hspace{1mm}Zeqiu Xu$^{2}$} \\
	Information Networking Institute\\
	Carnegie Mellon University\\
	Pittsburgh, PA, 15213 \\
	\texttt{zeqiux@alumni.cmu.edu} \\
}
\else
\usepackage{authblk}

\newbox{\orcid}\sbox{\orcid}{\includegraphics[scale=0.06]{orcid.pdf}} 
\author[1]{%
	{\hspace{1mm}Jingyuan Yi\thanks{Corresponding author: \texttt{jingyuay@alumni.cmu.edu}}}%
}
\author[1]{%
	{\hspace{1mm}Peiyang Yu\thanks{Corresponding author: \texttt{peiyangy@alumni.cmu.edu}}}%
}
\author[2]{%
	{\hspace{1mm}Tianyi Huang\thanks{\texttt{tianyihuang@berkeley.edu}}}%
}
\author[2]{%
	{\hspace{1mm}Zeqiu Xu\thanks{Corresponding author: \texttt{zeqiux@alumni.cmu.edu}}}%
}

\affil[1]{Carnegie Mellon University}
\affil[2]{University of California, Berkeley}
\fi

\hypersetup{
pdftitle={Optimization of Transformer heart disease prediction model based on particle swarm optimization algorithm},
pdfsubject={q-bio.NC, q-bio.QM},
pdfauthor={Peiyang Yu, Jingyuan Yi, Tianyi Huang, Zeqiu Xu},
pdfkeywords={Particle swarm optimization, Transformer, Prediction of heart disease},
}

\begin{document}
	
\maketitle
\begin{abstract}
	Aiming at the latest particle swarm optimization algorithm, this paper proposes an improved Transformer model to improve the accuracy of heart disease prediction and provide a new algorithm idea based on particle swarm optimization (PSO). We first use three mainstream machine learning classification algorithms - decision tree, random forest and XGBoost, and then output the confusion matrix of these three models. The results showed that the random forest model had the best performance in predicting the classification of heart disease, with an accuracy of 92.2\%. Then, we apply the Transformer model based on PSO algorithm to the same dataset for the classification experiment. The results show that the classification accuracy of the model is as high as 96.5\%, 4.3\% higher than that of random forest, which verifies the effectiveness of PSO in optimizing Transformer model. The above research shows that PSO significantly improves Transformer performance in heart disease prediction. Improving the ability to predict heart disease is a global priority with benefits for all humankind. Accurate prediction can enhance public health, optimize medical resources, and reduce healthcare costs, leading to a healthier society. This advancement paves the way for more efficient health management and supports the foundation of a healthier, more resilient global community.
\end{abstract}

% keywords can be removed

\twocolumn
\section{Introduction}
As one of the major global health threats, heart disease has become an important cause of death. According to the statistics of the World Health Organization, cardiovascular diseases cause millions of deaths every year, which not only brings a significant impact on the quality of life of patients but also brings a huge economic burden to society and the medical system~\cite{b1}. Therefore, early prediction and intervention of heart disease are particularly important. With the continuous progress of medical technology, especially in the field of data collection and information processing, the prediction and diagnosis of heart disease have gradually changed from traditional manual experience to data-driven methods, which provides new possibilities for reducing the morbidity and mortality of heart disease~\cite{b2}.

The application of machine learning algorithms has shown great potential in healthcare domain such as skin cancer tissue classification~\cite{b3}, predicting heart disease, etc. Traditional heart disease detection methods often rely on the subjective judgment and experience of doctors; however, this is often affected by various factors, resulting in inadequate diagnostic accuracy~\cite{b4}. By analyzing large amounts of patient data, machine learning algorithms can identify underlying patterns and features to make more accurate predictions. By combining a variety of indicators related to heart disease (such as age, blood pressure, cholesterol levels, etc.), machine learning models can effectively assess an individual's risk of heart disease. At the same time, with the development of more complex algorithms e.g. deep learning, and more explainable methods e.g. AIX360~\cite{b5}, the predictive power and interpretable power of the model have also been significantly improved, allowing doctors to better understand the risk factors that lead to heart disease~\cite{b6}.

Moreover, the implications of machine learning in predicting heart disease go far beyond improving the accuracy of predictions. Its efficient data processing capabilities allow doctors to obtain diagnoses faster, enabling them to formulate interventions in the shortest possible time. In addition, machine learning can also improve the performance of the model by constantly learning and updating new data, which means it is able to adapt to the changing medical environment and emerging risk factors. In conclusion, the introduction of machine learning algorithms has brought revolutionary changes to the prediction and management of heart disease, indicating great potential in the future medical field~\cite{b7}.

\section{Data from data analysis}
\label{sec2}
The data set used in this paper is an open-source data set, which contains various physiological information of patients, totaling 1888 pieces of information. Each message recorded a patient's age, sex, type of chest pain, resting blood pressure, serum cholesterol level, fasting blood glucose, resting electrocardiogram results, maximum heart rate achieved, exercise-induced angina pectoris, peak exercise ST segment slopes, number of major vessels stained by X-ray, type of thalassemia, and final predictors (0 indicates no heart disease, 1 indicates heart disease), we selected some indicators and some data sets for display~\cite{b8}, as shown in Table \ref{t1}.
\begin{table}[!ht]
	\caption{Selected data sets}
	\centering
	\label{t1}
    
	\setlength{\tabcolsep}{3pt} 
	\begin{tabular}{llllllll}
		\toprule
		age & sex & trestbps & chol & fbs & thalachh & oldpeak & target \\
		\midrule
		63 & 1 & 145 & 233 & 1 & 150 & 2.3 & 1 \\
		
		37 & 1 & 130 & 250 & 0 & 187 & 3.5 & 1 \\
		
		41 & 0 & 130 & 204 & 0 & 172 & 1.4 & 1 \\
		
		56 & 1 & 120 & 236 & 0 & 178 & 0.8 & 1 \\
		
		57 & 0 & 120 & 354 & 0 & 163 & 0.6 & 1 \\
		
		57 & 1 & 140 & 192 & 0 & 148 & 0.4 & 1 \\
		
		56 & 0 & 140 & 294 & 0 & 153 & 1.3 & 1 \\
		
		44 & 1 & 120 & 263 & 0 & 173 & 0 & 1 \\
		
		52 & 1 & 172 & 199 & 1 & 162 & 0.5 & 1 \\
		
		57 & 1 & 150 & 168 & 0 & 174 & 1.6 & 1 \\
		
		54 & 1 & 140 & 239 & 0 & 160 & 1.2 & 1 \\
		
		48 & 0 & 130 & 275 & 0 & 139 & 0.2 & 1 \\
		\bottomrule
	\end{tabular}
\end{table}

"target" is the predictor variable, and other variables are the input variables. 1 in target indicates yes, and 2 indicates no.

\section{Correlation analysis}
\begin{figure}[!h]
	\centering
	\includegraphics[width=.9\linewidth]{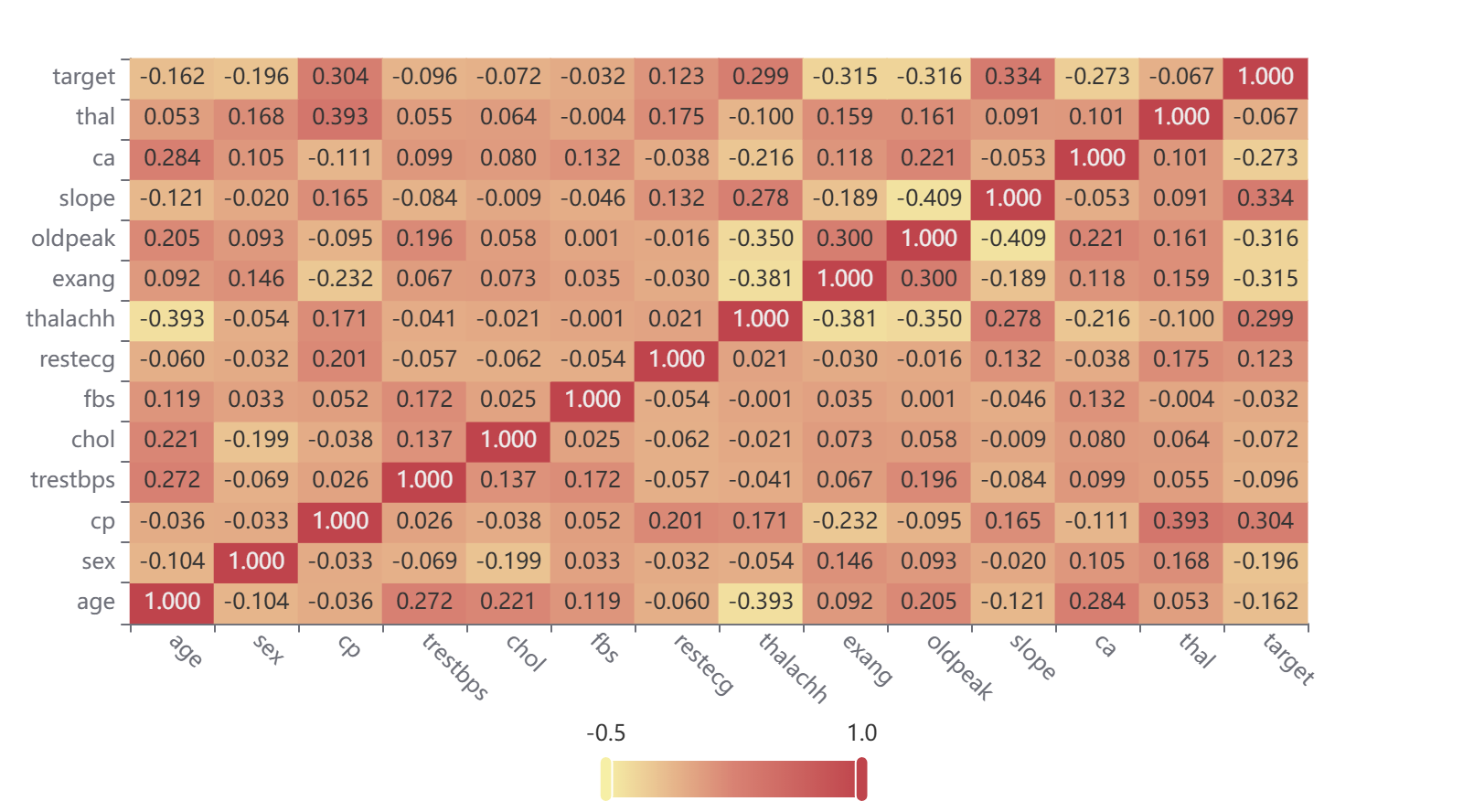}
	\caption{Correlation heat maps.}
	\label{f1}
\end{figure}

Correlation analysis is used to assess the strength and direction of linear relationships between two or more variables. The most common correlation measure is the Pearson correlation coefficient, which ranges from -1 to 1. Correlation analysis was carried out on various variables in the data, and correlation heat maps were drawn, as shown in Figure \ref{f1}.

For the analysis of this correlation heat map, we can see that the values in the figure range from -1 to 1, indicating the strength of the correlation between different variables. The color depth of the thermal map can intuitively indicate the intensity of the correlation. The darker the color, the stronger the correlation. As can be seen from the figure, there is a relatively strong correlation between some variables, which can be used for machine learning analysis and prediction.

\section{Method}

\subsection{Related works}
In the domain of heart disease prediction, various machine learning (ML) methods have been employed to enhance the accuracy and efficiency of diagnosis. Notably, the application of Particle Swarm Optimization (PSO) algorithm has demonstrated significant improvement in optimizing machine learning models. For instance, studies have integrated PSO with stacked sparse auto-encoders to achieve a classification accuracy of 96.1\%, outperforming several state-of-the-art methods~\cite{b9}. This approach is particularly valuable due to its efficiency in learning from limited samples, which is crucial given the complexity and cost associated with data collection in clinical settings. Furthermore, PSO has been utilized to optimize the parameters of Support Vector Machines(SVM), leading to effective performance in various classification and prediction problems, including heart disease prediction. The combination of PSO with SVM has resulted in hybrid models that outperform conventional ones, highlighting the importance of PSO in adjusting model parameters to achieve optimal results~\cite{b10}.

The selection of the PSO algorithm for optimizing the Transformer-based heart disease prediction model is driven by several compelling reasons.Firstly,PSO's ability to efficiently handle complex optimization tasks with few parameters makes it an attractive choice for adjusting the parameters of deep learning models like Transformers.Its simplicity in implementation and effectiveness in balancing exploration and exploitation phases are particularly beneficial for enhancing model performance.Secondly,PSO's stochastic nature allows it to navigate the search space effectively,which is crucial for identifying optimal solutions in the context of heart disease prediction where the dataset can be multimodal and complex.Lastly,the integration of PSO with Transformers leverages the strengths of both approaches: PSO's optimization capabilities and Transformers'ability to capture long-range dependencies and produce context-aware representations.This synergy is expected to yield a model that not only optimizes prediction accuracy but also maintains computational efficiency,especially when dealing with large datasets.

\subsection{Particle swarm optimization}
\begin{figure}[!h]
	\centering
	\includegraphics[width=.5\linewidth]{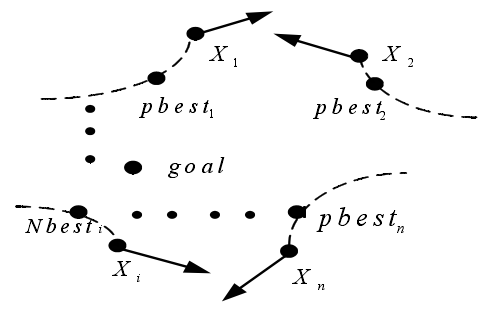}
	\caption{The schematic diagram of particle swarm optimization.}
	\label{f2}
\end{figure}

Particle swarm optimization (PSO) is an optimization algorithm based on swarm intelligence. PSO is inspired by the collective behavior of birds when searching for food, and its basic principle is to find the optimal solution by simulating the movement and mutual cooperation of a group of particles in the search space. These particles move through the multidimensional search space, each particle representing a potential solution, and the state of the particle is determined by two parameters: its position and velocity. The motion of particles is affected by their own historical optimal position and the historical optimal position of the whole group, thus achieving global optimization~\cite{b11}. The schematic diagram of particle swarm optimization is shown in Figure \ref{f2}.

In the PSO algorithm, a fitness function is first defined to evaluate the performance of each particle~\cite{b12}. Each particle is randomly initialized with position and velocity in the search space, and then its fitness value is calculated. As the iteration progresses, each particle gradually approaches the optimal solution by updating its own speed and position. In terms of velocity update, the current velocity of a particle is composed of three components: the current velocity, the attraction of the particle's own historical optimal location, and the attraction of the group's historical optimal location. Specifically, when updating the speed of a particle, it takes into account the best place it has been (individual experience) and the best place it has found in the entire particle population (social experience), thus guiding the search process with information from both aspects.

In the process of updating the motion of the particle, the position and velocity of the particle are adjusted simultaneously. The new speed is updated according to the formula to control the amplitude of movement in different directions. The formula usually contains several parameters, such as learning factor, inertia weight, etc., which can regulate the exploration and development ability of particles. The inertia weight determines the search range of the particle in the search process. The larger inertia weight is conducive to global search, while the smaller inertia weight is conducive to local optimization. By adjusting these parameters, the algorithm can effectively balance the relationship between exploring the unknown region and using the known information, so as to improve the convergence speed and search efficiency.

\subsection{Transformer}
\begin{figure}[!h]
	\centering
	\includegraphics[width=.6\linewidth]{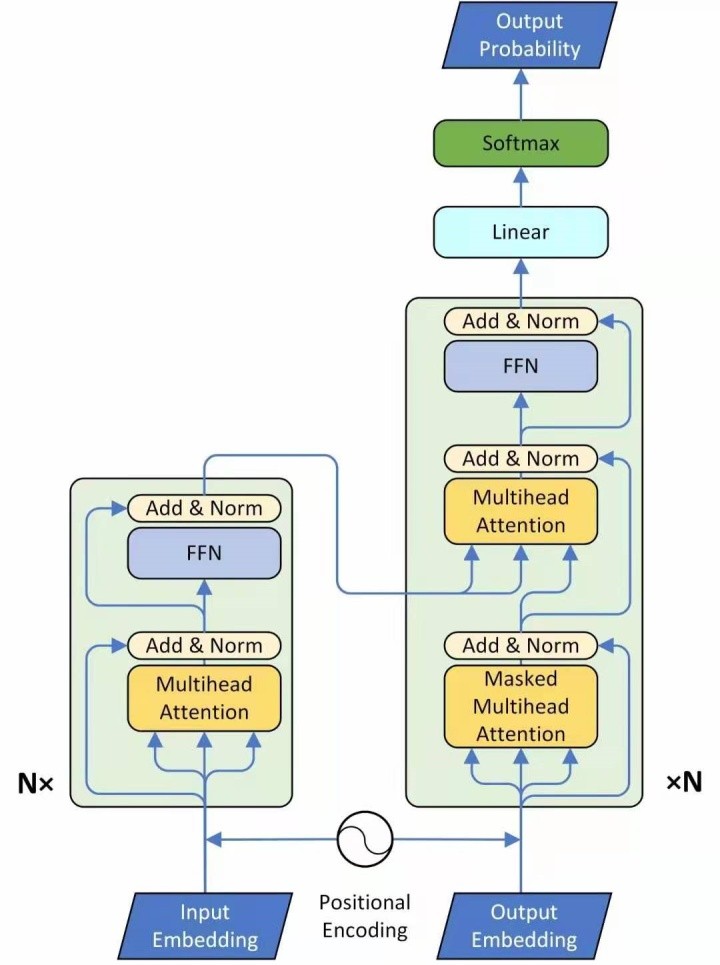}
	\caption{The schematic diagram of Transformer.}
	\label{f3}
\end{figure}

Transformer is a deep learning model used to process sequential data. Different from traditional recurrent neural networks (RNN) and long short-term memory (LSTM) networks, Transformer is completely based on self-attention mechanism and does not rely on sequential processing of sequences. It can process data in parallel and improve training efficiency, especially when processing long sequences~\cite{b9}. The schematic diagram of Transformer is shown in Figure \ref{f3}.

Transformer architecture is mainly divided into Encoder and Decoder two parts. The function of the encoder is to encode the input sequence into a representation vector that contains contextual information. Each encoder layer consists of two main components: a self-attention mechanism and a fully connected feedforward neural network. The self-attention mechanism allows the model to dynamically consider the effects of other words in the sequence while processing the current word, generating a weighted representation for each word by calculating the relevance of each word to the others (attention weights). Fully connected feedforward neural networks are responsible for further processing of these weighted representations. In encoders, the superposition of multiple layers can enhance the expressiveness of the model.

The structure of the decoder is similar to that of the encoder, but it requires additional consideration of the context information provided by the encoder when generating the output sequence. The self-attention mechanism in the decoder is designed to mask self-attention to ensure that only current and previous words can be considered during the generation process, and no future information can be revealed. This ensures that the generated sequence is in order. At each layer of the decoder, there are also fully connected feedforward neural networks and attentional mechanisms for the encoder output, through which the decoder is able to make full use of the context information generated by the encoder to produce a more coherent and context-relevant output.

Another core innovation of Transformers is Positional Encoding, since the self-attention mechanism itself does not contain positional information for the sequence. In order for the model to recognize the position information of elements in the sequence, position coding is added to the input embeddings so that the model can effectively understand the relative position relationship of individual words in the sequence. This design makes Transformer better able to handle long text, while avoiding the common gradient disappearance and explosion problems of RNN and LSTM in long sequence processing.

\subsection{Improved transformer model based on particle swarm optimization algorithm}
The improved Transformer model based on particle swarm optimization algorithm aims to enhance the performance of Transformer in specific tasks by using the optimization capability of PSO. In the Transformer model optimized based on PSO, PSO is mainly used to optimize Transformer hyperparameters and model structure. In this paper, PSO is used to optimize the learning rate, number of layers, hidden layer dimensions and number of attention heads of Transformer model. The first is the learning rate. Different tasks may require different learning rates, and PSO can effectively explore the optimal learning rate. The second is the number of layers and hidden layer dimensions, and the structural parameters of the model can be optimized by particles to find the best network depth and width, thereby improving the performance of the model. The last is the number of attention heads. In multi-head self-attention mechanism, optimizing the number of attention heads can better capture context information.

\begin{figure}[!h]
	\centering
	\includegraphics[width=.8\linewidth]{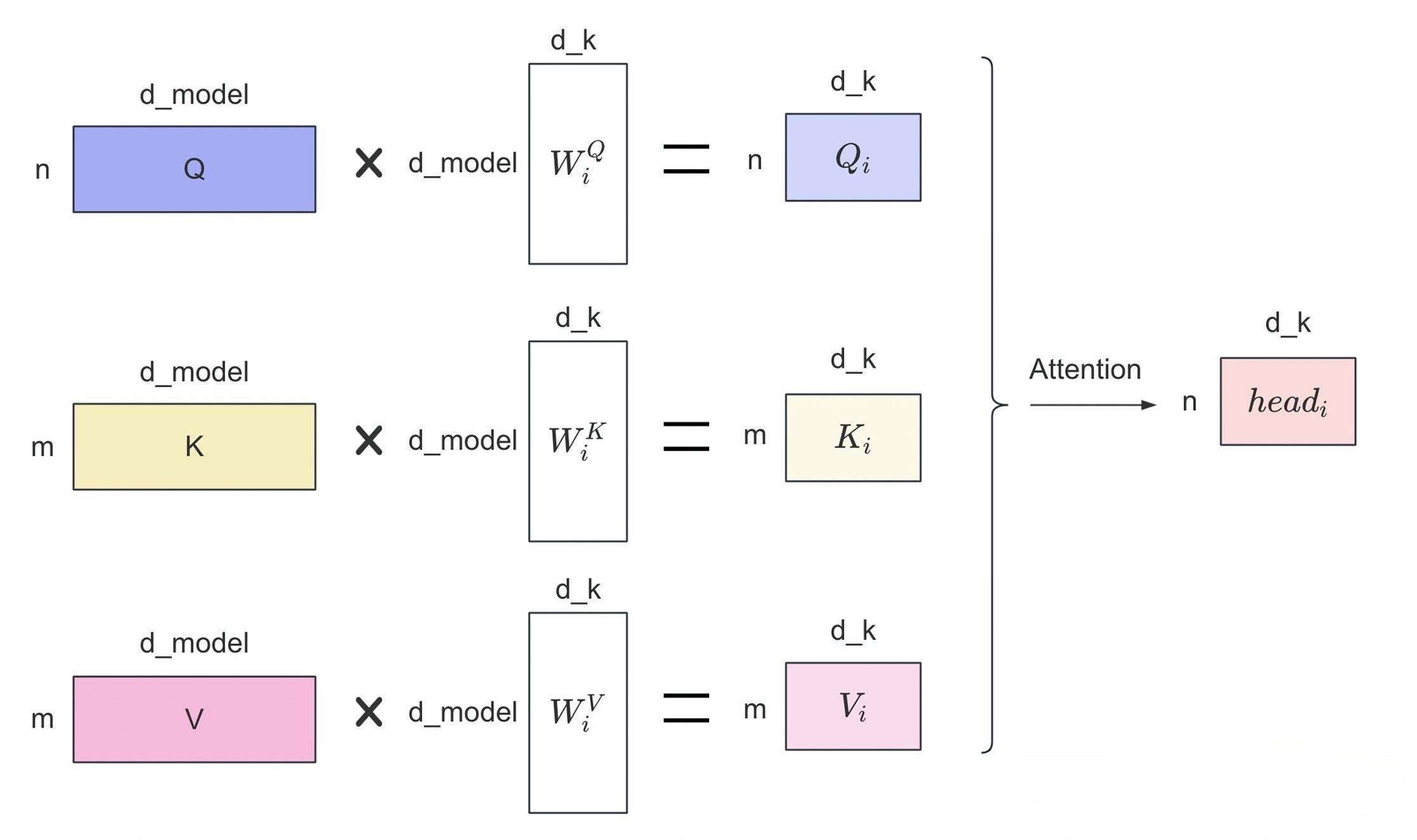}
	\caption{The optimization process.}
	\label{f4}
\end{figure}

In essence, PSO treats the hyperparameters of the Transformer model as a multidimensional search space. The model first randomly initializes the particle's position (i.e. the hyperparameter combination) and velocity. These combinations are then used to train the Transformer model and evaluate its performance on specific tasks (e.g., accuracy, loss, etc.) to determine fitness values. The particle then updates its speed and position to explore more optimized hyperparameters. Through many iterations, the PSO can find the optimal combination of hyperparameters. The optimization process is shown in Figure \ref{f4}.

In the experiment of Transformer classification algorithm based on particle swarm optimization (PSO) algorithm, the hardware configuration is 3060 graphics card and the memory is 32G.

In terms of model parameter settings, the learning rate of Transformer model is set to 0.001, the batch size is 32, the optimizer selects Adam, the training rounds are 50, the hidden layer dimension is 512, the number of attention heads is 8, and the hidden layer dimension of feedforward network is 2048. In addition, the number of particles in the PSO algorithm is set to 30, the maximum number of iterations is set to 100, the initial value of inertia weight w is set to 0.9, and the individual and group experience weights c1 and c2 are set to 2.

\begin{figure}[!h]
	\centering
	\includegraphics[width=.6\linewidth]{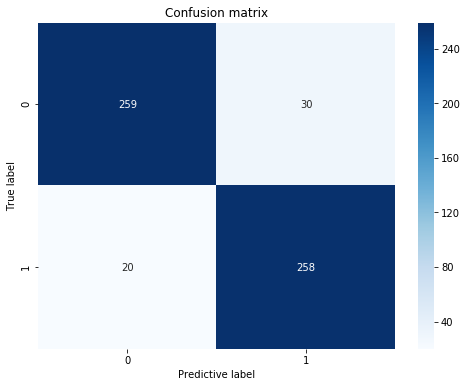}
	\caption{The confusion matrix of the decision tree model.}
	\label{f5}
\end{figure}

\begin{figure}[!h]
	\centering
	\includegraphics[width=.6\linewidth]{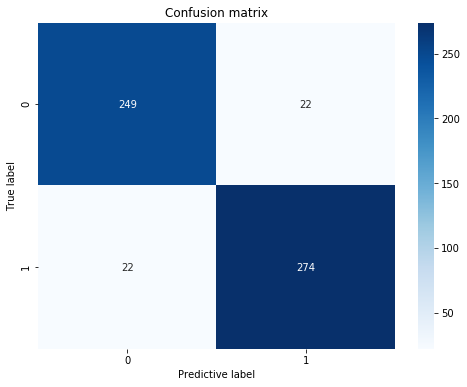}
	\caption{The confusion matrix of the random forest model.}
	\label{f6}
\end{figure}

\begin{figure}[!h]
	\centering
	\includegraphics[width=.6\linewidth]{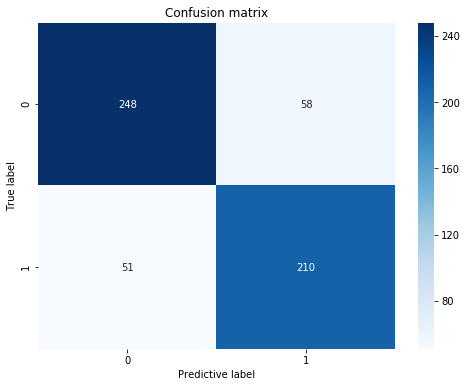}
	\caption{The confusion matrix of theXGBoost model.}
	\label{f7}
\end{figure}

\begin{figure}[!h]
	\centering
	\includegraphics[width=.6\linewidth]{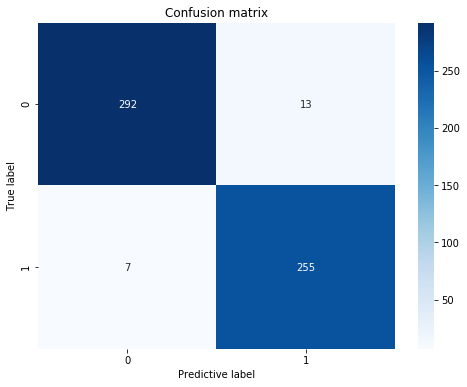}
	\caption{The confusion matrix of our model.}
	\label{f8}
\end{figure}

Firstly, three commonly used machine learning algorithms (decision tree, random forest and XGBoost) are used for classification experiments, and the confusion matrix of the three models is output. The confusion matrix of the decision tree model test set is shown in Figure \ref{f5}, the random forest model test set is shown in Figure \ref{f6}, and the XGBoost model test set is shown in Figure \ref{f7}. Finally, Transformer based on particle swarm optimization (PSO) algorithm in this paper is used for classification experiment, and the output confusion matrix is shown in Figure \ref{f8}.

Output three basic machine learning models and model evaluation indicators of the model in this paper. In this paper, accuracy, precision, recall rate and F1 score are used to evaluate the model, and the results are shown in Table \ref{t2}.

\begin{table}[!h]
	\caption{Model evaluation}
	\centering
	\label{t2}
	\begin{tabular}{llllllll}
		\toprule
		& Accuracy  & Precision & recall  & F1 \\
		\midrule
		Decision tree & 0.912 & 0.896 & 0.928 & 0.911 \\
		
		Random forest & 0.922 & 0.926 & 0.926 & 0.926 \\
		
		XGBoost & 0.808 & 0.784 & 0.805 & 0.794 \\
		
		Textual model & 0.965 & 0.952 & 0.973 & 0.962 \\

		\bottomrule
	\end{tabular}

\end{table}

\begin{figure}[!h]
	\centering
	\includegraphics[width=.6\linewidth]{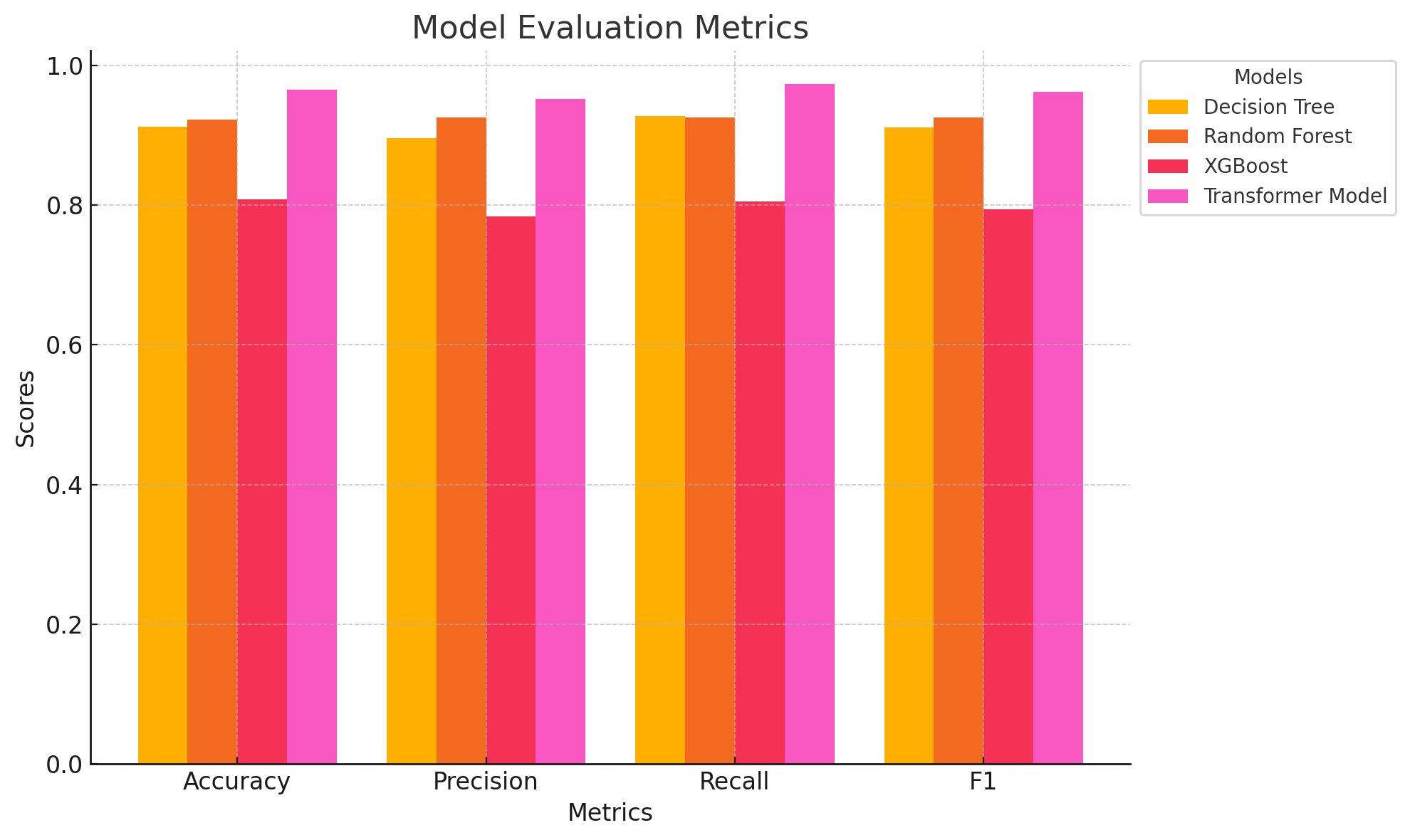}
	\caption{The comparison of model parameters.}
	\label{f9}
\end{figure}

The comparison of model parameters is shown in Figure \ref{f9}.

Experimental results show that among the three basic models, random forest model has the best effect in the classification and prediction of heart disease, with an accuracy rate of 92.2\%; while the classification accuracy rate of Transformer model optimized based on particle swarm optimization in this paper reaches 96.5\%, 4.3\% higher than that of random forest, effectively improving the accuracy rate of heart disease prediction.

\section{Conclusion}
In this paper, the latest particle swarm optimization (PSO) is used to optimize Transformer model to improve the effect of heart disease prediction, and a new algorithm is proposed. By combining the optimization capabilities of PSO with the Transformer model, the research aims to enhance the model's performance in specific medical tasks. In the experiment, we used three classical machine learning algorithms - decision tree, random forest and XGBoost - to make categorical predictions of heart disease and output their confusion matrix respectively. The results show that the random forest model has the best performance among the three, with an accuracy of 92.2\%. However, the PSO optimized Transformer model has achieved a significant improvement in classification accuracy, reaching 96.5\%, which is 4.3\% higher than random forest, proving the effectiveness of the proposed method.

It can be seen from the experimental results that the Transformer model optimized based on particle swarm optimization not only improves the ability to predict heart disease, but also shows its application potential in the medical field. Heart disease is a critical health issue impacting millions globally. Enhancing the accuracy of heart disease prediction holds immense value, benefiting individuals and reducing healthcare costs. Advanced algorithms and technologies promise to improve clinical outcomes, ease the strain on public health systems, and promote a healthier, more resilient global society.

\bibliographystyle{plainnat}    % 支持作者-年份格式的样式文件
\bibliography{references}  

\begin{thebibliography}{12}
\providecommand{\natexlab}[1]{#1}
\providecommand{\url}[1]{\texttt{#1}}
\expandafter\ifx\csname urlstyle\endcsname\relax
  \providecommand{\doi}[1]{doi: #1}\else
  \providecommand{\doi}{doi: \begingroup \urlstyle{rm}\Url}\fi

\bibitem[Arya et~al.(2021)Arya, Bellamy, Chen, Dhurandhar, Hind, Hoffman, Houde, Liao, Luss, Mojsilovi\'{c}, Mourad, Pedemonte, Raghavendra, Richards, Sattigeri, Shanmugam, Singh, Varshney, Wei, and Zhang]{b5}
Vijay Arya, Rachel K.~E. Bellamy, Pin-Yu Chen, Amit Dhurandhar, Michael Hind, Samuel~C. Hoffman, Stephanie Houde, Q.~Vera Liao, Ronny Luss, Aleksandra Mojsilovi\'{c}, Sami Mourad, Pablo Pedemonte, Ramya Raghavendra, John Richards, Prasanna Sattigeri, Karthikeyan Shanmugam, Moninder Singh, Kush~R. Varshney, Dennis Wei, and Yunfeng Zhang.
\newblock Ai explainability 360 toolkit.
\newblock In \emph{Proceedings of the 3rd ACM India Joint International Conference on Data Science \& Management of Data (8th ACM IKDD CODS \& 26th COMAD)}, CODS-COMAD '21, page 376–379, New York, NY, USA, 2021. Association for Computing Machinery.
\newblock ISBN 9781450388177.
\newblock \doi{10.1145/3430984.3430987}.
\newblock URL \url{https://doi.org/10.1145/3430984.3430987}.

\bibitem[Doe et~al.(2024)Doe, Smith, Brown, and Lee]{b4}
John Doe, Alice Smith, Richard Brown, and Michael Lee.
\newblock Heart disease diagnosis using deep learning.
\newblock In \emph{IEEE Conference on Biomedical Systems}, volume~7, pages 120--128. IEEE, 2024.

\bibitem[Freitas et~al.(2020)Freitas, Lopes, and Morgado-Dias]{b11}
Diogo Freitas, Luiz~Guerreiro Lopes, and Fernando Morgado-Dias.
\newblock Particle swarm optimisation: A historical review up to the current developments.
\newblock \emph{Entropy}, 22\penalty0 (3):\penalty0 362, 2020.
\newblock \doi{10.3390/e22030362}.

\bibitem[Hassan et~al.(2004)Hassan, Cohanim, de~Weck, and Venter]{b12}
Rania Hassan, Babak Cohanim, Olivier de~Weck, and Gerhard Venter.
\newblock A comparison of particle swarm optimization and the genetic algorithm.
\newblock \emph{MIT Engineering Systems Division Working Paper Series}, pages 1--30, 2004.
\newblock URL \url{http://dspace.mit.edu/handle/1721.1/6018}.

\bibitem[Hossain et~al.(2024)Hossain, Talukder, and Mahmud]{b2}
Md.~Sahadat Hossain, Md.~Alamin Talukder, and Md.~Zulfiker Mahmud.
\newblock Advancements in cardiovascular disease detection: Leveraging data mining and machine learning.
\newblock \emph{bioRxiv}, 2024.
\newblock \doi{10.1101/2024.03.09.584222}.

\bibitem[Hussein and Abdulazeez(2024)]{b3}
R.~Hussein and A.~Abdulazeez.
\newblock Skin cancer detection utilizing deep learning: Classification of skin lesion images using a vision transformer.
\newblock \emph{Journal of Applied Science and Technology Trends}, 5\penalty0 (2):\penalty0 60--71, 2024.
\newblock Discusses deep learning algorithms like CNNs and their applications in distinguishing between benign and malignant skin lesions.

\bibitem[Kazemnejad(2020)]{b10}
Amirhossein Kazemnejad.
\newblock Understanding self-attention and positional encoding of the transformer architecture.
\newblock \emph{Personal Blog}, 2020.
\newblock URL \url{https://kazemnejad.com/blog/transformer_architecture_positional_encoding/}.

\bibitem[Organization(2024)]{b1}
World~Health Organization.
\newblock Global cardiovascular disease statistics and impacts: Key findings and challenges.
\newblock \emph{World Health Statistics Report}, 2024.
\newblock Available at \url{https://www.who.int}.

\bibitem[Patro and Padhy(2023)]{b6}
Sibo~Prasad Patro and Neelamadhab Padhy.
\newblock A secure remote health monitoring for heart disease prediction using machine learning and deep learning techniques in explainable artificial intelligence framework.
\newblock In \emph{Proceedings of the 10th International Electronic Conference on Sensors and Applications (ECSA-10)}, pages 78--86. MDPI, 2023.
\newblock URL \url{https://doi.org/10.3390/ecsa-10-16237}.

\bibitem[Sengupta et~al.(2019)Sengupta, Basak, and Peters]{b7}
Saptarshi Sengupta, Sanchita Basak, and Richard Alan~II Peters.
\newblock Particle swarm optimization: A survey of historical and recent developments with hybridization perspectives.
\newblock \emph{Machine Learning and Knowledge Extraction}, 1\penalty0 (1):\penalty0 157--191, 2019.
\newblock \doi{10.3390/make1010010}.

\bibitem[Vaswani et~al.(2017{\natexlab{a}})Vaswani, Shazeer, Parmar, Uszkoreit, Jones, Gomez, Kaiser, and Polosukhin]{b8}
Ashish Vaswani, Noam Shazeer, Niki Parmar, Jakob Uszkoreit, Llion Jones, Aidan~N. Gomez, Łukasz Kaiser, and Illia Polosukhin.
\newblock Attention is all you need.
\newblock \emph{Advances in Neural Information Processing Systems}, 30:\penalty0 5998--6008, 2017{\natexlab{a}}.

\bibitem[Vaswani et~al.(2017{\natexlab{b}})Vaswani, Shazeer, Parmar, Uszkoreit, Jones, Gomez, Kaiser, and Polosukhin]{b9}
Ashish Vaswani, Noam Shazeer, Niki Parmar, Jakob Uszkoreit, Llion Jones, Aidan~N Gomez, Łukasz Kaiser, and Illia Polosukhin.
\newblock Attention is all you need.
\newblock \emph{Advances in Neural Information Processing Systems}, 30:\penalty0 5998--6008, 2017{\natexlab{b}}.

\end{thebibliography}

\end{document}